\documentclass[compactaffiliation]{Interspeech}

\interspeechcameraready

\title{A Multi-Dialectal Dataset for German Dialect ASR\\and Dialect-to-Standard Speech Translation}

\author[affiliation={1,2}]{Verena}{Blaschke}
\author[affiliation={1}]{Miriam}{Winkler}
\author[affiliation={3}]{Constantin}{Förster}
\author[affiliation={3}]{Gabriele}{Wenger-Glemser}
\author[affiliation={1,2}]{Barbara}{Plank}

\affiliation{MaiNLP, CIS}{LMU Munich}{Germany}
\affiliation{}{Munich Center for Machine Learning}{Germany}
\affiliation{}{Bayerischer Rundfunk}{Germany}
\email{\{verena.blaschke, b.plank\}@lmu.de, \{constantin.foerster, gabriele.wenger-glemser\}@br.de}


\usepackage{xcolor}

\usepackage{booktabs}
\usepackage{multirow}
\usepackage{adjustbox}

\usepackage{tikz}
\usetikzlibrary{calc}
\usetikzlibrary{matrix,shapes,arrows,fit,tikzmark}

\usepackage{pgfplots, pgfplotstable}
\usetikzlibrary{shapes.geometric, }
\pgfplotsset{compat=1.18}
\usepgfplotslibrary{groupplots}

\newcommand{\betthupferl}{Betthupferl}

\usepackage{fontawesome5}

\colorlet{blueLight}{blue!30!white}
\colorlet{blueMid}{blue!55!white}
\colorlet{blueDark}{blue!50!gray}
\colorlet{redLight}{red!30!white}
\colorlet{redMid}{red!55!white}
\colorlet{redDark}{red!50!gray}
\colorlet{orangeLight}{orange!30!white}
\colorlet{orangeMid}{orange!55!white}
\colorlet{orangeDark}{orange!50!gray}

\newcommand{\bluecircleLight}{\textcolor{blueLight}{\faCircle}}
\newcommand{\bluecircleMid}{\textcolor{blueMid}{\faCircle}}
\newcommand{\bluecircleDark}{\textcolor{blueDark}{\faCircle}}
\newcommand{\redtriangleLight}{\textcolor{redLight}{$\blacktriangle$}}
\newcommand{\redtriangleMid}{\textcolor{redMid}{$\blacktriangle$}}
\newcommand{\redtriangleDark}{\textcolor{redDark}{$\blacktriangle$}}
\newcommand{\orangesquareLight}{\textcolor{orangeLight}{\faSquare}}
\newcommand{\orangesquareMid}{\textcolor{orangeMid}{\faSquare}}
\newcommand{\orangesquareDark}{\textcolor{orangeDark}{\faSquare}}

\newcommand{\bluetiny}{{\scriptsize\bluecircleDark\bluecircleMid\bluecircleLight}}
\newcommand{\redtiny}{{\redtriangleDark\kern-0.3pt\redtriangleMid\kern-0.3pt\redtriangleLight}}
\newcommand{\orangetiny}{{\scriptsize\orangesquareDark\kern0.7pt\orangesquareMid\kern0.7pt\orangesquareLight}}

\newcommand{\sameAsRef}{\faCheckCircle}
\newcommand{\validAlt}{\faCheckCircle[regular]}
\newcommand{\error}{\faTimes}

\usepackage{stfloats} %
\usepackage[bottom]{footmisc}

\begin{document}

\maketitle

\begin{abstract}
Although Germany has a diverse landscape of dialects, they are underrepresented in current automatic speech recognition (ASR) research.
To enable studies of how robust models are towards dialectal variation, 
we present \betthupferl{}, an 
evaluation dataset containing four hours of read speech in three dialect groups spoken in Southeast Germany (Franconian, Bavarian, Alemannic), and half an hour of Standard German speech. 
We provide both dialectal and Standard German transcriptions, and analyze the linguistic differences between them. 
We benchmark several multilingual state-of-the-art ASR models on speech translation into Standard German, and find differences between how much the output resembles the dialectal vs.\ standardized transcriptions.
Qualitative error analyses of the best ASR model reveal that it sometimes normalizes grammatical differences, but often stays closer to the dialectal constructions.
\end{abstract}

\section{Introduction}

Although non-standard dialects are widely spoken, so far they have received less focus in speech technology research than standard languages.
Current automatic speech recognition (ASR) systems perform worse on non-standard dialects (cf.\ \cite{tatman2017effects, markl2022language} and \S\ref{sec:related-work}).
At the same time, 
speakers of German dialects have expressed interest in ASR systems that support their dialects, with Standard German and/or dialectal output~\cite{blaschke-etal-2024-dialect}.

A challenge for automatically transcribing dialectal audio is that differences between standard languages and non-standard dialects do not only include pronunciation differences (like in ASR for non-standard accents~\cite{aksenova2022accented}), but also lexical, morphological, and syntactic differences.
Prior work on ASR for dialectal data typically does not analyze
the -- valid or incorrect -- ASR transcription differences related to these linguistic differences.

To enable more research on this, we release \betthupferl{},\footnote{%
We release the transcriptions, annotations, annotation guidelines, data statement, code, and model predictions under \texttt{github.com/mainlp/betthupferl}.
Access to the audio data must be granted by Bayerischer Rundfunk on a case-by-case basis due to copyright restrictions (contact Gabriele Wenger-Glemser).} a dataset containing audios from three underrepresented German dialect groups with dialectal and Standard German reference transcriptions~(\S\ref{sec:dataset}).
It is the first dataset for Franconian, Bavarian, and Swabian ASR evaluation.
We benchmark several state-of-the-art ASR systems on this dataset~(\S\ref{sec:experiments}), and validate automatic ASR quality metrics with human judgments~(\S\ref{sec:human-judgments}).
Furthermore, we analyze the differences between the dialectal and Standard German transcriptions~(\S\ref{sec:data-differences}) and build on this in our ASR error analysis~(\S\ref{sec:error-analysis}).

\begin{figure}[t]
    \centering
    \includegraphics[width=0.8\columnwidth]{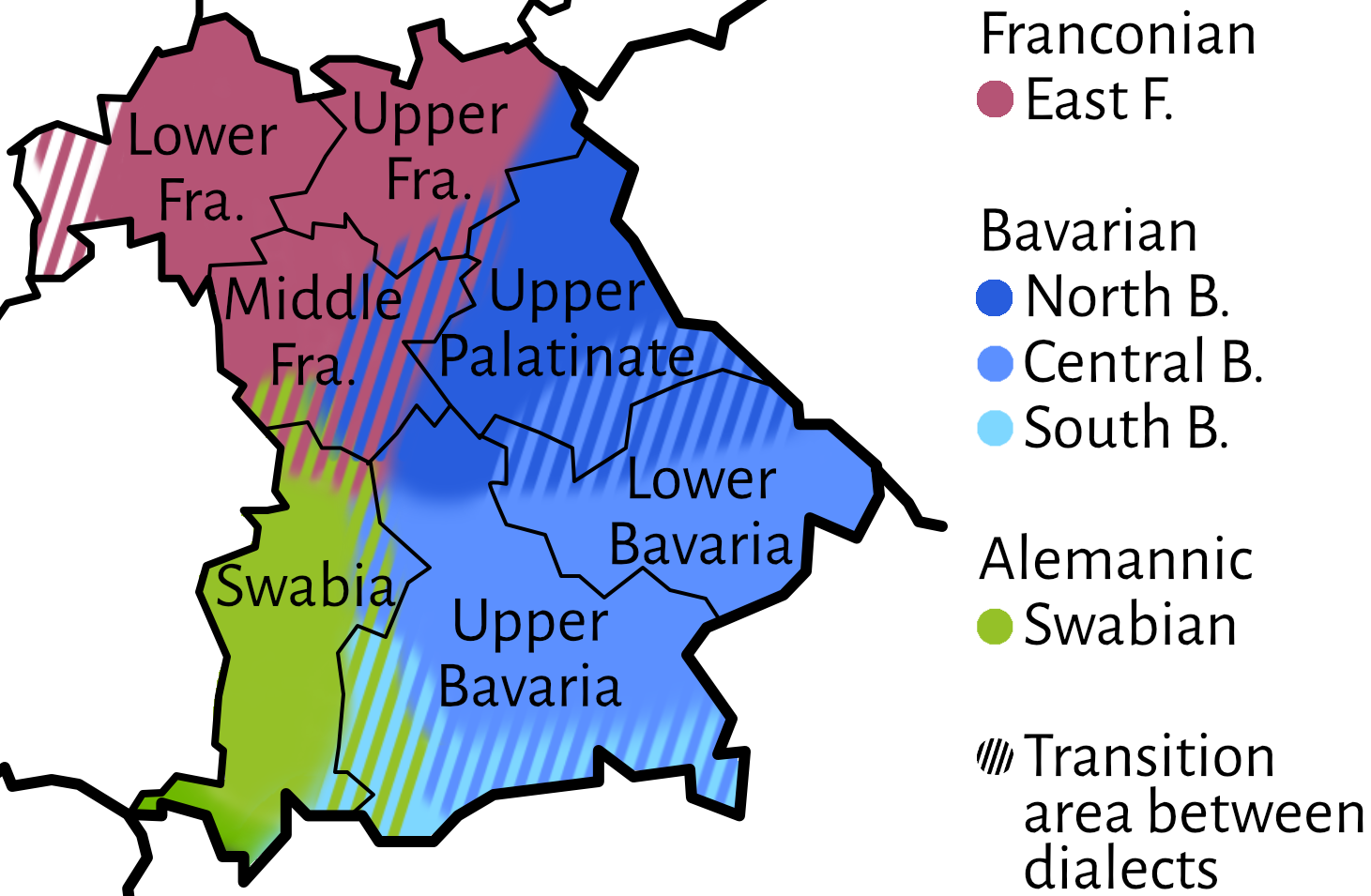}
    \caption{The dialects and administrative regions included in \betthupferl{} (dialect division after \cite{wiesinger1983einteilung}).
    }
    \label{fig:map}
\end{figure}

Figure~\ref{fig:map} shows the dialects in our dataset, and Table~\ref{tab:example} shows an example of our data.

\section{Related work}
\label{sec:related-work}

ASR for non-standard dialects has received increased focus in the last few years.
A major line of research has focused on ASR producing standard language output (including most of the works cited here), 
while another line has (also or instead) focused on preserving variation~\cite{ahia-etal-2024-voices, markl2023automatic, nigmatulina-etal-2020-asr}.

Works focusing on automatically transcribing dialectal German data are the most relevant to our work.
\cite{blaschke-etal-2023-survey} covers a summary of German dialect speech datasets. 
Most relevantly, this includes two Swiss German datasets that are transcribed both in Standard German and in the dialect \cite{scherrer2019archimob, dogans2021swissdial}.
Furthermore, we note a study on (similarly to us) using public broadcast data to build a speech dataset for Upper Saxon~\cite{herms-etal-2016-corpus}, and works on datasets including many (Swiss) German dialects \cite{pluss-etal-2022-sds, pluss-etal-2023-stt4sg}.
Moreover, some recent works analyze the performance of Whisper on Swiss German data~\cite{dolev-etal-2024-whisper, sicard-etal-2023-spaiche}.
Additional work focuses on the post-correction of literally transcribed dialectal data to match the word choice and grammar of Standard German~\cite{gerlach-etal-2022-producing}.

{\tikzset{   
        every picture/.style={remember picture,baseline},
        every node/.style={anchor=base,align=center,outer sep=1.5pt},
        every path/.style={thick},
        }

\newcommand\detnameTopLeft{%
    \tikz[overlay, remember picture] 
        \node (marker-detname-a) at (.5em, .2em) {};%
}
\newcommand\detnameBottomRight{%
    \tikz[overlay, remember picture] 
        \node (marker-detname-b) at (-0.25em, .3em) {};%
    \tikz[overlay, remember picture, inner sep=2pt]
        \node[draw=teal!80, dashed, rounded corners, fit=(marker-detname-a.north west) (marker-detname-b.south east)] {};%
}
\newcommand\possTopLeft{%
    \tikz[overlay, remember picture] 
        \node (marker-poss-a) at (.5em, .5em) {};%
}
\newcommand\possBottomRight{%
    \tikz[overlay, remember picture] 
        \node (marker-poss-b) at (-.5em, .6em) {};%
    \tikz[overlay, remember picture, inner sep=2pt]
        \node[draw=orange, dotted, line width=1.3pt, rounded corners, fit=(marker-poss-a.north west) (marker-poss-b.south east)] {};%
}
\newcommand\lexiconTopLeft{%
    \tikz[overlay, remember picture] 
        \node (marker-lexicon-a) at (.5em, .3em) {};%
}
\newcommand\lexiconBottomRight{%
    \tikz[overlay, remember picture] 
        \node (marker-lexicon-b) at (-0.7em, .5em) {};%
    \tikz[overlay, remember picture, inner sep=2pt]
        \node[draw=purple!60, dashed, rounded corners, fit=(marker-lexicon-a.north west) (marker-lexicon-b.south east)] {};%
}
\newcommand{\diffnote}[2]{}

\begin{table*}[t]
\caption{Example sentence from the `Middle Franconia' split.
In addition to phonetic and morphological differences that are reflected in the spelling, the references differ in \textcolor{teal}{whether a determiner is added before the name}, in the \textcolor{orange}{possessive construction}, and in a \textcolor{purple}{choice of phrase}.
The hypothesis generated by Whisper large-v3 (hypo) sometimes matches the Standard German (Std) reference~{\footnotesize\sameAsRef}, sometimes follows the dialectal version (dial) in a way that is acceptable in (informal) German~{\footnotesize\validAlt},
and sometimes contains errors~\error\kern1pt.
}
\label{tab:example}
\newcommand{\gloss}[1]{\textit{\textcolor{gray}{#1}}}
\setlength{\tabcolsep}{2pt}
\adjustbox{max width=\textwidth}{%
\begin{tabular}{@{}llllllllllllllllllll@{}}
& \multicolumn{17}{@{}l}{\textit{``Everybody, immediately spread out and search for Mathilda's coin, or I'll show you what's what!''}} & \\[3pt]
\textbf{Std} & Sofort & alle & ausswärmen & und & \detnameTopLeft & \possTopLeft{}Mathildas &  & Geldstück & suchen, & sonst & zeige & ich & euch, & \multicolumn{2}{l}{\lexiconTopLeft{}wo's} & \multicolumn{2}{l}{langgeht.} & \multicolumn{2}{r}{\diffnote{purple}{word choice}} \\
& \gloss{At once} & \gloss{all} & \gloss{spread out} & \gloss{and} &  & \multicolumn{2}{l}{\gloss{Mathilda.\textsc{gen}}}  & \gloss{coin} & \gloss{search} & \gloss{else} & \gloss{show} & \gloss{I} & \gloss{you} & \multicolumn{2}{l}{\gloss{where it}} & \multicolumn{4}{l}{\gloss{runs along.}} \\
\textbf{Dial} & Sofort & alle & ausschwärma & und & da & Mathilda & ihr & Geldstückle & sung, & sonst & zach & ich & eich, & wo & da & Bartl & an & Most & hoid. \\
& &  & \multicolumn{2}{r}{\diffnote{teal}{det.\ + name}} & \gloss{the} & \gloss{Mathilda}\detnameBottomRight & \gloss{her}\possBottomRight & \multicolumn{3}{l}{\diffnote{orange}{poss.}} & &  &  & \gloss{where} & \gloss{the} & \gloss{Barthel} & \gloss{the} & \gloss{cider} & \gloss{fetches}.\lexiconBottomRight \\
\textbf{Hypo} & Sofort & alle & Ausschwärmer & und & der & Mathilda & ihr & Geldstück & lesung. & Sonst & zeig & ich & euch, & wo & der & Badl & den & Most & holt. \\
& \multicolumn{1}{c}{\sameAsRef} & \multicolumn{1}{c}{\sameAsRef} & \multicolumn{1}{c}{\error{} \gloss{swarmers}} & \multicolumn{1}{c}{\sameAsRef} & \multicolumn{1}{c}{\validAlt} & \multicolumn{1}{c}{\sameAsRef} & \multicolumn{1}{c}{\validAlt} & \multicolumn{1}{c}{\sameAsRef} & \llap{\error{} }\gloss{reading} & \multicolumn{1}{c}{\sameAsRef} & \multicolumn{1}{c}{\validAlt} & \multicolumn{1}{c}{\sameAsRef} & \multicolumn{1}{c}{\sameAsRef} & {\validAlt} & \multicolumn{1}{c}{\validAlt} & ~~\error & \multicolumn{1}{c}{\validAlt} & \multicolumn{1}{c}{\validAlt} & ~{\validAlt} \\
\end{tabular}
}
\end{table*}
}

\begin{table}
\caption{%
Overview of \betthupferl, including mean sentence lengths for dialectal (dial) and Standard German (Std) references, and mean word-/character-level Levenshtein distances between them (in~\%).
Standard deviations are in subscripts. 
}
\label{tab:data-splits}
\begin{adjustbox}{max width=\columnwidth}
\begin{tabular}{@{}l@{}r@{\hspace{5pt}}r@{\hspace{5pt}}r@{\hspace{7pt}}r@{\hspace{3pt}}r@{\hspace{2pt}}r@{\hspace{4pt}}r@{}}
\toprule
 &  &  &  & \multicolumn{2}{c}{\textbf{Words/sent}} & \multicolumn{2}{@{}c@{}}{\textbf{Lev dist}} \\ 
 \cmidrule(lr){5-6} \cmidrule(lr){7-8}
\textbf{Region/split} & \textbf{\llap{S}peakers} & {\textbf{Sent}} & {\textbf{Min}} & \multicolumn{1}{c}{\textbf{Dial}} & \multicolumn{1}{c}{\textbf{Std}} & \textbf{Word} & \textbf{Char}\\ \midrule
Lower Franconia & 1F, 1M & 403 & 33 & 12.6\textsubscript{7.6} & 12.5\textsubscript{7.6} & 46\textsubscript{21} & 19\textsubscript{11} \\
Upper Franconia & 3M & 561 & 33 & 9.1\textsubscript{5.7} & 9.0\textsubscript{5.6} & 52\textsubscript{22} & 23\textsubscript{13} \\
Middle Franconia & 4M & 371 & 36 & 15.1\textsubscript{8.9} & 15.2\textsubscript{8.8} & 57\textsubscript{20} & 23\textsubscript{11} \\
Upper Palatinate & 1F, 1M & 394 & 34 & 14.0\textsubscript{9.0} & 13.9\textsubscript{8.9} & 58\textsubscript{19} & 24\textsubscript{11} \\
Lower Bavaria & 2F, 1M & 488 & 32 & 10.8\textsubscript{7.3} & 11.1\textsubscript{7.4} & 68\textsubscript{21} & 30\textsubscript{12} \\
Upper Bavaria & 1F, 2M & 465 & 37 & 11.8\textsubscript{8.0} & 12.1\textsubscript{8.2} & 57\textsubscript{21} & 23\textsubscript{11} \\
Swabia & 1F, 1M & 575 & 37 & 10.5\textsubscript{6.6} & 10.7\textsubscript{6.7} & 57\textsubscript{22} & 22\textsubscript{12} \\
\midrule
All dialects & 6F, 13M & 3\,257 & 241 & 11.7\textsubscript{7.7} & 11.8\textsubscript{7.8} & 57\textsubscript{22} & 24\textsubscript{12} \\
\midrule
\midrule
Std. German & 6F, 7M & 531 & 32 & {---}\phantom{---} & 8.9\textsubscript{5.6} & {---}\phantom{--} & {---}\phantom{--} \\
\midrule
Full dataset & 8F, 14M & 3\,788 & 273 & ---\phantom{---}  & 11.4\textsubscript{7.6} & ---\phantom{--}  & ---\phantom{--}  \\
\bottomrule
\end{tabular}
\end{adjustbox}
\end{table}

\section{\betthupferl}
\label{sec:dataset}

\subsection{Audio data}

\betthupferl{} contains four hours of read dialectal speech and half an hour of read Standard German speech (Table~\ref{tab:data-splits}).
It consists of recordings of stories for young children originally 
aired by the broadcaster Bayerischer Rundfunk, who
gave us permission to use their audio data. 
The stories take \qtyrange[range-units=single,range-phrase=--]{3}{4}{\minute} each and were created by authors with backgrounds in children's literature, screenplays, theatre, and radio. 
They were recorded by professional speakers selected by an editorial team for their ability to speak the native dialect of their respective regions in Bavaria. 
Each instance is one sentence (split manually during transcription), with a mean duration of 4.3$\pm$2.9\,s. %
The dataset contains \qtyrange[range-units=single,range-phrase=--]{32}{37}{\minute} of dialectal data from each of the seven administrative regions of the German state of Bavaria (Figure~\ref{fig:map}). 
This includes dialects from three dialect groups: Franconian, Bavarian, and Alemannic.
While the data are not parallel across varieties, they share the same genre and style.

In addition, the series contains stories recorded in Standard German, from which we include a sample in \betthupferl.
Some of the dialectal stories include characters that speak Standard German instead of the local dialect -- we include sentences from their dialogue in the Standard German test set instead of the respective dialectal test set.
Except for the Standard German excerpts from the otherwise dialectal audios, no speaker appears in more than one region or data split.

To our knowledge, there are currently no public data sets of the Franconian, Bavarian, or Swabian dialects suitable for automatic speech recognition, so we resorted to using copyrighted data obtained from a public broadcaster.

\subsection{Transcription}
\label{sec:transcription}

We provide two transcriptions for each dialectal audio: a dialectal transcription and a Standard German translation. Table~\ref{tab:example} contains an example.
We include the dialectal version in order to analyze how the Standard German version differs from the original audio data~(\S\ref{sec:data-differences}) and to what extent these differences are challenging to ASR systems~(\S\ref{sec:error-analysis}).
The Standard German audios are only transcribed in Standard German.

Because none of the dialects included in our dataset have any widely used orthography, we use ad-hoc pronunciation spellings based on Standard German grapheme--phoneme mappings, as those are widely used in natural language processing (NLP) for dialects \cite{blaschke-etal-2023-survey}.
The Standard German translations are close to the dialectal transcriptions, but can deviate due to lexical or grammatical differences.
The references were transcribed/\allowbreak{}translated by an in-house trained annotator, who
was hired and compensated 
following national salary rates. %
The transcriber is a native speaker of both Standard German and a Central Bavarian dialect from Upper Bavaria with a background in NLP, and used FOLKER %
\cite{schmidt-schutte-2010-folker} as transcription tool. %
Transcription %
amounts to \qtyrange[range-units=single,range-phrase=--]{5}{8}{\minute} per minute of dialectal data and \qtyrange[range-units=single,range-phrase=--]{3}{5}{\minute} per minute of Standard German audio.

\subsection{Differences between the transcriptions}
\label{sec:data-differences}

\begin{table}
\caption{%
\textbf{Sent/word (\S\ref{sec:data-differences}):}~Proportion of Standard German sentences/\allowbreak{}words containing at least one kind of difference to their dialectal counterparts (more than one kind can apply simultaneously).
\textbf{Hypothesis (\S\ref{sec:error-analysis}):}~Proportion of words in Whisper large-v3's hypotheses that are identical to the Standard German reference~({\footnotesize\sameAsRef}), different but valid~({\footnotesize\validAlt}), or errors~({\error}).
}
\label{tab:wordlevel-differences}
\centering
\adjustbox{max width=\columnwidth}{%
\begin{tabular}{@{}lrrrrrr@{}}
\toprule
\multicolumn{3}{r}{Proportion w. type of diff. (\%)} & \multicolumn{4}{@{}r@{}}{Hypothesis words} \\ 
\cmidrule(lr){2-3} \cmidrule(l){5-7} 
Difference & Sent & \llap{W}ord && {\sameAsRef} & {\validAlt} & {\error} \\
\midrule
--- (identical word) & {97} & {45} && 86 & 4 & 10 \\
Phonetic/morphological & {96} & {47} && 75 & 5 & 20 \\
Word splitting & {41} & {4} && 54 & 10 & 36 \\
Determiner + name & {29} & {3} && 10 & 77 & 13 \\
Word choice & {23} & {2} && 8 & 30 & 63 \\
Verb phrase construction & {7} & {1} && 13 & 23 & 63 \\
Word order & {6} & {1} && 0 & 82 & 18 \\
Dropped/fused pronoun & {5} & {0} && 40 & 0 & 60 \\
Possessive & {2} & {0} && 0 & 57 & 43 \\
Other & {8} & {1} && 27 & 47 & 27 \\ 
\bottomrule
\end{tabular}}
\end{table}

To quantify how similar the dialectal and Standard German references are to each other, we calculate the Levenshtein distances between each pair of references and normalize it by the length of the longer sequence.\footnote{%
For comparability with the word/character error rates~(\S\ref{sec:quant-analysis}), we ignore casing and punctuation. 
We split words based on whitespace.
}
The mean word-level distance per sentence is \qty{57}{\percent}; the character-level distance is \qty{24}{\percent} (Table~\ref{tab:data-splits}).

As a qualitative comparison, we annotate word-level differences on a subset of the dataset (one story per region, totalling 4.2k words in 305 sentences) based on nine categories (see Table~\ref{tab:wordlevel-differences}).
The annotator is a native speaker of German and has a background in linguistics.
We annotate spelling differences (reflecting phonetic and/or morphological differences),
different word (or phrase) choices, as well as a range of grammatical differences that have been the focus of prior German dialect NLP work~\cite{gerlach-etal-2022-producing, artemova-etal-2024-exploring, blaschke-etal-2024-maibaam}.

The first three columns of Table~\ref{tab:wordlevel-differences} summarize the differences.
Nearly half the words are identical in both transcript versions (and almost every sentence contains at least one identically spelled word).
Slightly more have spelling differences that reflect phonetic and/or morphological differences.
Relatively few differences are due to differently split/fused words (\qty{4}{\percent} -- although \qty{41}{\percent} of sentences include at least one such word splitting difference), lexical differences (\qty{2}{\percent}), word order differences (\qty{1}{\percent}), or other grammatical differences (\qty{5}{\percent} total).

\section{Experiments}
\label{sec:experiments}

We select several ASR models~(\S\ref{sec:models}) for a quantitative evaluation on Betthupferl~(\S\ref{sec:quant-analysis}).
We further report human judgments~(\S\ref{sec:human-judgments}) and an error analysis~(\S\ref{sec:error-analysis}) of the best model's ASR hypotheses.

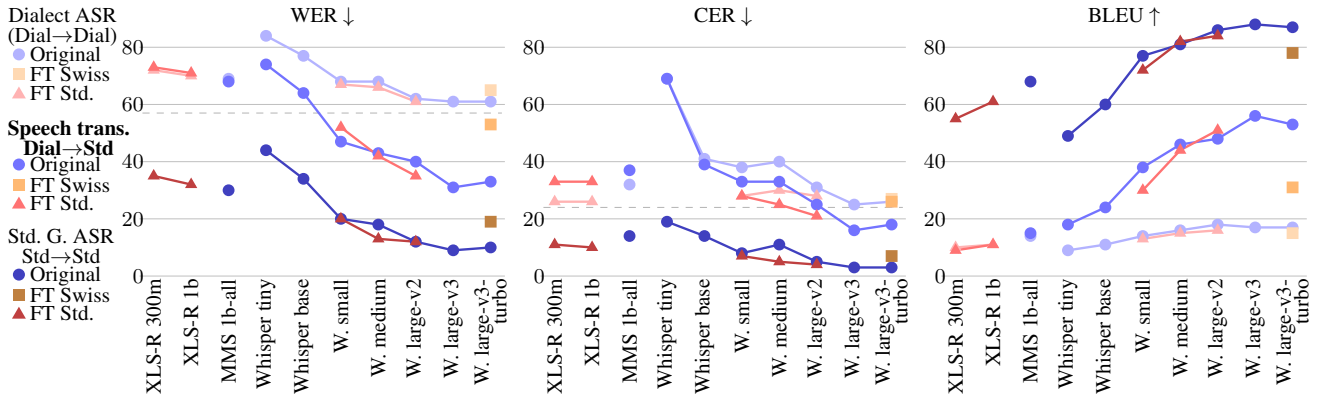
\begin{figure*}[t]
    \centering
    \begin{adjustbox}{max width=\textwidth, trim=9pt 6pt 4pt 6pt}%
    \pgfplotstableread[header=true]{
Model	dummy
XLS-R~300m	0
XLS-R~1b	0
MMS~1b-all	0
Whisper~tiny	0	
Whisper~base	0
W.~small	0
W.~medium	0
W.~large-v2	0	
W.~large-v3	0
{W.~large-v3-\\[-3pt]turbo}	0
}\alldata
\pgfplotstableread[header=true]{
Model	WER-dial-dial	WER-dial-deu	WER-deu-deu	CER-dial-dial	CER-dial-deu	CER-deu-deu	BLEU-dial-dial	BLEU-dial-deu	BLEU-deu-deu
tiny	84	74	44	69	69	19	9	18	49
base	77	64	34	41	39	14	11	24	60
small	68	47	20	38	33	8	14	38	77
medium	68	43	18	40	33	11	16	46	81
large-v2	62	40	12	31	25	5	18	48	86
large-v3	61	31	9	25	16	3	17	56	88
large-v3-turbo	61	33	10	26	18	3	17	53	87
}\vanilladata
\pgfplotstableread[header=true]{
Model	WER-dial-dial	WER-dial-deu	WER-deu-deu	CER-dial-dial	CER-dial-deu	CER-deu-deu	BLEU-dial-dial	BLEU-dial-deu	BLEU-deu-deu
mms-1b-all	69	68	30	32	37	14	14	15	68
}\mmsdata
\pgfplotstableread[header=true]{
Model	WER-dial-dial	WER-dial-deu	WER-deu-deu	CER-dial-dial	CER-dial-deu	CER-deu-deu	BLEU-dial-dial	BLEU-dial-deu	BLEU-deu-deu
small-german	67	52	20	28	28	7	13	30	72
medium-german	66	42	13	30	25	5	15	44	82
large-v2-german	61	35	12	28	21	4	16	51	84
}\germanwhisper
\pgfplotstableread[header=true]{
Model	WER-dial-dial	WER-dial-deu	WER-deu-deu	CER-dial-dial	CER-dial-deu	CER-deu-deu	BLEU-dial-dial	BLEU-dial-deu	BLEU-deu-deu
xls-r-300m-german	72	73	35	26	33	11	10	9	55
xls-r-1b-german	70	71	32	26	33	10	11	11	61
}\germanxlsr
\pgfplotstableread[header=true]{
Model	WER-dial-dial	WER-dial-deu	WER-deu-deu	CER-dial-dial	CER-dial-deu	CER-deu-deu	BLEU-dial-dial	BLEU-dial-deu	BLEU-deu-deu
large-v3-turbo-swissgerman	65	53	19	27	26	7	15	31	78
}\gswft
\begin{tikzpicture}
\begin{groupplot}[
   group style={group size=3 by 1, horizontal sep=15pt},
   xtick=data,
   xticklabels from table={\alldata}{Model},
   xticklabel style={rotate=90},
   ymajorgrids,
   y axis line style={opacity=0},
   tick style={draw=none},
   enlarge x limits={abs=6pt},
   clip=false, %
   height=5cm,
   yticklabel style={xshift=5pt},
   xticklabel style={xshift=2pt, align=right},
]

\nextgroupplot[
 width=7cm,
 title=WER $\downarrow$,
   ymin=0, ymax=80,
   ytick={0, 20, 40, 60, 80},
]

\node[label={[anchor=170, align=center]:Dialect ASR\\[-2pt](Dial$\rightarrow$Dial)}] at (-4.1, 88.5){};
\node[circle, fill=blueLight, inner sep=1.8pt, label={[anchor=170]:\,Original}] at (-3.6, 77.5){};
\node[rectangle, fill=orangeLight, inner sep=2.4pt, label={[anchor=173]:\,FT Swiss}] at (-3.6, 70.5){};
\node[regular polygon, regular polygon sides=3, fill=redLight, inner sep=1.2pt, label={[anchor=170]:\,FT Std.}] at (-3.6, 63.5){};

\node[label={[anchor=170, align=center]:\textbf{Speech trans.}\\[-2pt]\textbf{Dial$\rightarrow$Std}}] at (-4.1, 50){};
\node[circle, fill=blueMid, inner sep=1.8pt, label={[anchor=170]:\,Original}] at (-3.6, 39){};
\node[rectangle, fill=orangeMid, inner sep=2.4pt, label={[anchor=173]:\,FT Swiss}] at (-3.6, 32){};
\node[regular polygon, regular polygon sides=3, fill=redMid, inner sep=1.2pt, label={[anchor=170]:\,FT Std.}] at (-3.6, 25){};

\node[label={[anchor=170, align=center]:Std.\ G.\ ASR\\[-2pt]Std$\rightarrow$Std}] at (-4.1, 11.5){};
\node[circle, fill=blueDark, inner sep=1.8pt, label={[anchor=170]:\,Original}] at (-3.6, 0.5){};
\node[rectangle, fill=orangeDark, inner sep=2.4pt, label={[anchor=173]:\,FT Swiss}] at (-3.6, -6.5){};
\node[regular polygon, regular polygon sides=3, fill=redDark, inner sep=1.2pt, label={[anchor=170]:\,FT Std.}] at (-3.6, -13.5){};

\draw[dashed, gray!60] (axis cs:-0.3, 57) -- (axis cs:9.3, 57);

\addplot[draw=none] table [y=dummy, x expr=\coordindex,] {\alldata};
\addplot[blueDark, mark=*, line width=1pt,
] table [y=WER-deu-deu, x expr=\coordindex+3,] {\vanilladata};
\addplot[blueDark, mark=*, line width=1pt,] table [y=WER-deu-deu, x expr=\coordindex+2,] {\mmsdata};
\addplot[redDark, mark=triangle*, line width=1pt] table [y=WER-deu-deu, x expr=\coordindex+5,] {\germanwhisper};
\addplot[redDark, mark=triangle*, line width=1pt] table [y=WER-deu-deu, x expr=\coordindex,] {\germanxlsr};
\addplot[orangeDark, mark=square*, line width=1pt] table [y=WER-deu-deu, x expr=\coordindex+9,] {\gswft};

\addplot[blueLight, mark=*, line width=1pt] table [y=WER-dial-dial, x expr=\coordindex+3,] {\vanilladata};
\addplot[blueLight, mark=*, line width=1pt,] table [y=WER-dial-dial, x expr=\coordindex+2,] {\mmsdata};
\addplot[redLight, mark=triangle*, line width=1pt] table [y=WER-dial-dial, x expr=\coordindex+5,] {\germanwhisper};
\addplot[redLight, mark=triangle*, line width=1pt] table [y=WER-dial-dial, x expr=\coordindex,] {\germanxlsr};
\addplot[orangeLight, mark=square*, line width=1pt] table [y=WER-dial-dial, x expr=\coordindex+9,] {\gswft};

\addplot[blueMid, mark=*, line width=1pt] table [y=WER-dial-deu, x expr=\coordindex+3,] {\vanilladata};
\addplot[blueMid, mark=*, line width=1pt,] table [y=WER-dial-deu, x expr=\coordindex+2,] {\mmsdata};
\addplot[redMid, mark=triangle*, line width=1pt] table [y=WER-dial-deu, x expr=\coordindex+5,] {\germanwhisper};
\addplot[redMid, mark=triangle*, line width=1pt] table [y=WER-dial-deu, x expr=\coordindex,] {\germanxlsr};
\addplot[orangeMid, mark=square*, line width=1pt] table [y=WER-dial-deu, x expr=\coordindex+9,] {\gswft};

\nextgroupplot[
 width=7cm,
 title=CER $\downarrow$,
   ymin=0, ymax=80,
   ytick={0, 20, 40, 60, 80},
]
\draw[dashed, gray!60] (axis cs:-0.3, 24) -- (axis cs:9.3, 24);

\addplot[draw=none] table [y=dummy, x expr=\coordindex,] {\alldata};
\addplot[blueDark, mark=*, line width=1pt] table [y=CER-deu-deu, x expr=\coordindex+3,] {\vanilladata};
\addplot[blueDark, mark=*, line width=1pt,] table [y=CER-deu-deu, x expr=\coordindex+2,] {\mmsdata};
\addplot[redDark, mark=triangle*, line width=1pt] table [y=CER-deu-deu, x expr=\coordindex+5,] {\germanwhisper};
\addplot[redDark, mark=triangle*, line width=1pt] table [y=CER-deu-deu, x expr=\coordindex,] {\germanxlsr};
\addplot[orangeDark, mark=square*, line width=1pt] table [y=CER-deu-deu, x expr=\coordindex+9,] {\gswft};

\addplot[blueLight, mark=*, line width=1pt] table [y=CER-dial-dial, x expr=\coordindex+3,] {\vanilladata};
\addplot[blueLight, mark=*, line width=1pt,] table [y=CER-dial-dial, x expr=\coordindex+2,] {\mmsdata};
\addplot[redLight, mark=triangle*, line width=1pt] table [y=CER-dial-dial, x expr=\coordindex+5,] {\germanwhisper};
\addplot[redLight, mark=triangle*, line width=1pt] table [y=CER-dial-dial, x expr=\coordindex,] {\germanxlsr};
\addplot[orangeLight, mark=square*, line width=1pt] table [y=CER-dial-dial, x expr=\coordindex+9,] {\gswft};

\addplot[blueMid, mark=*, line width=1pt] table [y=CER-dial-deu, x expr=\coordindex+3,] {\vanilladata};
\addplot[blueMid, mark=*, line width=1pt,] table [y=CER-dial-deu, x expr=\coordindex+2,] {\mmsdata};
\addplot[redMid, mark=triangle*, line width=1pt] table [y=CER-dial-deu, x expr=\coordindex+5,] {\germanwhisper};
\addplot[redMid, mark=triangle*, line width=1pt] table [y=CER-dial-deu, x expr=\coordindex,] {\germanxlsr};
\addplot[orangeMid, mark=square*, line width=1pt] table [y=CER-dial-deu, x expr=\coordindex+9,] {\gswft};

\nextgroupplot[
 width=7cm,
 title=BLEU $\uparrow$,
   ymin=0, ymax=80,
   ytick={0, 20, 40, 60, 80},
]
\addplot[draw=none] table [y=dummy, x expr=\coordindex,] {\alldata};
\addplot[blueDark, mark=*, line width=1pt] table [y=BLEU-deu-deu, x expr=\coordindex+3,] {\vanilladata};
\addplot[blueDark, mark=*, line width=1pt,] table [y=BLEU-deu-deu, x expr=\coordindex+2,] {\mmsdata};
\addplot[redDark, mark=triangle*, line width=1pt] table [y=BLEU-deu-deu, x expr=\coordindex+5,] {\germanwhisper};
\addplot[redDark, mark=triangle*, line width=1pt] table [y=BLEU-deu-deu, x expr=\coordindex,] {\germanxlsr};
\addplot[orangeDark, mark=square*, line width=1pt] table [y=BLEU-deu-deu, x expr=\coordindex+9,] {\gswft};

\addplot[blueLight, mark=*, line width=1pt] table [y=BLEU-dial-dial, x expr=\coordindex+3,] {\vanilladata};
\addplot[blueLight, mark=*, line width=1pt,] table [y=BLEU-dial-dial, x expr=\coordindex+2,] {\mmsdata};
\addplot[redLight, mark=triangle*, line width=1pt] table [y=BLEU-dial-dial, x expr=\coordindex+5,] {\germanwhisper};
\addplot[redLight, mark=triangle*, line width=1pt] table [y=BLEU-dial-dial, x expr=\coordindex,] {\germanxlsr};
\addplot[orangeLight, mark=square*, line width=1pt] table [y=BLEU-dial-dial, x expr=\coordindex+9,] {\gswft};

\addplot[blueMid, mark=*, line width=1pt] table [y=BLEU-dial-deu, x expr=\coordindex+3,] {\vanilladata};
\addplot[blueMid, mark=*, line width=1pt,] table [y=BLEU-dial-deu, x expr=\coordindex+2,] {\mmsdata};
\addplot[redMid, mark=triangle*, line width=1pt] table [y=BLEU-dial-deu, x expr=\coordindex+5,] {\germanwhisper};
\addplot[redMid, mark=triangle*, line width=1pt] table [y=BLEU-dial-deu, x expr=\coordindex,] {\germanxlsr};
\addplot[orangeMid, mark=square*, line width=1pt] table [y=BLEU-dial-deu, x expr=\coordindex+9,] {\gswft};

\end{groupplot}
\end{tikzpicture}%
    \end{adjustbox}%
    \caption{%
    Quantitative results for all models, audio/reference language variety combinations, and automatic metrics.
    \textbf{Models:} We compare the originally released multilingual ASR models 
    \mbox{({\footnotesize\bluecircleDark\bluecircleMid\bluecircleLight})}
    to versions fine-tuned on Standard German 
    \mbox{(\kern-1pt{\large\redtriangleDark\kern-0.5pt\redtriangleMid\kern-0.5pt\redtriangleLight})}
    and Swiss German \mbox{(\kern0.7pt{\footnotesize\orangesquareDark\kern0.5pt\orangesquareMid\kern0.5pt\orangesquareLight}).}
    Lines connect results that belong to models of the same family.
    \textbf{Language variety combinations:} We compare the results for dialect-to-Standard-German speech translation 
    \mbox{({\footnotesize\bluecircleMid}\kern-0.5pt{\large\redtriangleMid}{\footnotesize\orangesquareMid})}
    to those for dialect ASR 
    (light colours,
    \mbox{{\footnotesize\bluecircleLight}\kern-0.5pt{\large\redtriangleLight}{\footnotesize\orangesquareLight}),}
    and Standard German ASR 
    (dark colours,
    \mbox{{\footnotesize\bluecircleDark}\kern-0.5pt{\large\redtriangleDark}{\footnotesize\orangesquareDark}).}
    \textbf{Metrics:} We compare WER, CER, and BLEU. The dashed grey lines in the WER/CER plots indicate the mean normalized word-/character-level Levenshtein distances between the dialectal and German references.
    }
    \label{fig:wer-cer-bleu}
\end{figure*}

\begin{table}
\caption{%
The models we use.
\textbf{Licenses:} \textit{A:}~Apache~2.0, \textit{M:}~MIT, \textit{C:}~CC~BY-NC~4.0.
\textbf{Architecture:} \textit{C:}~convolution layers, \textit{E:}~encoders, \textit{D:}~decoders, \textit{LM:}~language modelling (via decoder layers), \textit{CTC:} connectionist temporal classification. 
}
\label{tab:models}
\adjustbox{max width=\columnwidth}{%
\begin{tabular}{@{}l@{\hspace{7pt}}l@{\hspace{7pt}}l@{\hspace{7pt}}l@{~}l@{\hspace{7pt}}l@{}r@{}}
\toprule
\textbf{Model} & \textbf{Size name} & \textbf{\llap{Li}c.} &
\multicolumn{2}{@{}l@{}}{\textbf{Fine-tuning}}
& \textbf{Layers} & \textbf{\llap{Dec}oding} \\ 
\midrule
Whisper & tiny & A & \bluetiny{} & none & 2C, 4E, 4D & LM  \\
 & base & A & \bluetiny{} & none & 2C, 6E, 6D & LM \\
 & small & A & \bluetiny{} & none & 2C, 12E, 12D & LM \\
 & small & A & \redtiny{} & Std.\ G & 2C, 12E, 12D & LM \\
 & medium & A & \bluetiny{} & none & 2C, 24E, 24D & LM \\
 & medium & A & \redtiny{} & Std.\ G & 2C, 24E, 24D & LM \\
 & large-v2 & A & \bluetiny{} & none & 2C, 32E, 32D & LM \\
 & large-v2 & A & \redtiny{} & Std.\ G & 2C, 32E, 32D & LM \\
 & large-v3 & A & \bluetiny{} & none & 2C, 32E, 32D & LM \\
 & large-v3-turbo & M & \bluetiny{} & none & 2C, 32E, 4D & LM \\
 & large-v3-turbo & M & \orangetiny{} & Swiss & 2C, 32E, 4D & LM \\
XLS-R & 300m & A & \redtiny{} & Std.\ G & 6C, 24E & CTC \\
 & 1b & A & \redtiny{} & Std.\ G & 6C, 48E & CTC \\
MMS & 1b-all & C & \bluetiny{} & none & 6C, 48E & CTC \\ 
\bottomrule
\end{tabular}
}
\end{table}

\subsection{Models}
\label{sec:models}

All models~(Table~\ref{tab:models}) were (pre-)trained on multilingual data (including Standard German data, but as far as we know no dialectal German data).
Where possible, we evaluate multiple available sizes per model to compare the effect of model size. %
If available, we include existing fine-tuned  models trained on (comparable) German data.
We evaluate several \textbf{Whisper} \cite{radford2023whisper} models.
Whisper has previously shown promising results for transcribing non-standard German data~\cite{dolev-etal-2024-whisper, schubert2024challenges, gorisch-schmidt-2024-evaluating}, but has also been shown to exhibit performance disparities when processing non-standard dialects~\cite{harris-etal-2024-modeling, kantharuban-etal-2023-quantifying}.
We additionally include versions of three Whisper sizes that are publicly available \cite{ardila-etal-2020-common} and were further fine-tuned on Standard German Common Voice data %
\cite{GermanWhisperCV11}.
We also experiment with one Whisper model that was fine-tuned on Swiss German dialects \cite{whisper-large-v3-turbo-swissgerman} (which are related to the Swabian dialect in our data) with a mix of Standard German and dialectal references.
Furthermore, we evaluate \textbf{XLS-R} \cite{babu2022xls-r} models in two sizes, fine-tuned on Standard German Common Voice data. %
\cite{wav2vec2-xls-r-300m-german-de, wav2vec2-xls-r-1B-german}.
Lastly, we include \textbf{MMS}~\cite{pratap2024mms}, 
as released for multilingual ASR.

Because none of the multilingual models explicitly support the dialects in our dataset, we set their input/output language to German. %
In our analysis, we focus on dialect-to-Standard-German speech translation (dialectal audio, Standard German reference),
but additionally provide scores for dialect ASR (same audio, but dialectal reference) and Standard German ASR (Standard German audio and reference) for comparison.

\subsection{Quantitative results and analysis}
\label{sec:quant-analysis}

We measure the word error rate (WER, in~\%) based on whitespace tokenization, the character error rate (CER, in~\%), and BLEU scores (via SacreBLEU 
v2.5.1~\cite{post-2018-call}).\footnote{Parameters: \texttt{nrefs:1|\allowbreak{}eff:no|\allowbreak{}tok:13a|\allowbreak{}smooth:exp}}
We ignore casing and punctuation, and report mean sentence-level scores.

Figure~\ref{fig:wer-cer-bleu} shows WER, CER, and BLEU scores.
Unsurprisingly, all models are better at transcribing the Standard German audio (dark colours) than the dialectal audio (medium and light colours).
Across all evaluation set-ups, Whisper large-v3 outperforms the other models.
Nevertheless, even for this model, a marked gap remains between its performance on producing Standard German transcriptions for dialectal vs.\ Standard German audio (WER: 31 vs.~9, 
CER: 16 vs.~3,
BLEU: 56 vs.~88).

Because of the limited amount of data per region or dialect group and the limited number of speakers, we focus on aggregate evaluations on the entire dialectal test set.
However, if we consider the de-aggregated results, we do not find any consistent patterns or strikingly different results regarding the region, dialect group, or speaker gender.

\textbf{The decoder type makes a difference.}
For the Whisper models (whose decoders act as a language model), the output for the dialectal audio is nearly always closer to the Standard German references than the dialectal ones.
When comparing the model hypotheses for dialectal audio to the dialectal references (light colours), the WER/CER appears to be bounded by the 
Levenshtein distance between the dialectal and Standard German references (dashed line). %
For the larger Whisper models, this appears to be due to the output being relatively standard-like (as evidenced by the low corresponding WER/CER when compared to the Standard German reference; medium colours).

For the connectionist temporal classification (CTC) models (MMS and XLS-R), %
the output is closer to the dialectal reference, as demonstrated by the CER scores (e.g., the output of the larger XLS-R model has a CER of~26 when compared to the dialectal references vs.\ 33 when compared to their German counterparts).
On a word level, the distances to Standard German and dialectal references are nearly identical.

\textbf{For Whisper as well as XLS-R, size matters:} the larger the model, the better the performance, regardless of the metric or audio/reference language combination.
The best overall performance is reached by Whisper large-v3.
However, its pruned version (large-v3-turbo) only shows a very minor performance decline despite having considerably fewer decoder layers.

\textbf{Fine-tuning on Standard and Swiss German data has mixed outcomes.}
Of the three Whisper models further fine-tuned on the Standard German Common Voice data, only the largest %
shows clear improvements over its original counterpart for dialect-to-Standard-German speech translation.
The smallest model performs worse on the dialectal data than its original counterpart; it might have overfit to its Standard German fine-tuning data.
The model fine-tuned on Swiss German (Alemannic) produces overall worse results than its original counterpart. %
While the performance degrades for nearly all dialects and whenever Standard German references are used, it improves marginally for dialect ASR on the only fellow Alemannic dialect in our dataset, Swabian (--1~WER, --6~CER, +1~BLEU).

\subsection{Qualitative analysis}
\label{sec:qual-analysis}

\subsubsection{Human judgments of ASR quality}
\label{sec:human-judgments}

To evaluate how well the automatic metrics align with human judgments, we annotate a subset of the outputs produced by the best ASR system, Whisper large-v3, based on whether they preserve the \textit{meaning} of the utterance and whether they are \textit{fluent/grammatical}, inspired by the evaluations in~\cite{dolev-etal-2024-whisper}.
Both dimensions are evaluated on a scale from 1~(worst) to 5~(best).
All three annotators are native speakers of German.
One also speaks East Franconian, another South Bavarian.
We selected 612 sentences (two stories per region) for annotation; two annotators annotated all of them, one a subset of 246 sentences.

\begin{table}
\caption{%
Human judgments of Whisper large-v3:
mean scores (meaning, fluency, 0.5$\times$(meaning+fluency)), 
inter-annotator agreement %
as mean pairwise correlations between annotators, 
and correlations with automatic metrics.
Correlations are Spearman's~$\rho$ ($p<0.001$). 
Std.\ deviations are in subscripts.
}
\label{tab:human-eval-correlations}
\adjustbox{max width=\columnwidth}{%
\setlength{\tabcolsep}{3pt}
\begin{tabular}{@{}lrrrrrr@{}}
\toprule
& \multicolumn{1}{c}{\multirow{2}{*}{Avg}} & \multicolumn{1}{c}{\multirow{2}{*}{IAA}} & \multicolumn{4}{r}{Correlations ($\rho$, mean over annotators)} \\ \cmidrule(l){4-7} 
\multicolumn{2}{r}{} &  & Fluency & WER & CER & BLEU \\
 \midrule
Meaning & {3.9}\textsubscript{1.1} & {0.76}\textsubscript{0.05} & {0.73}\textsubscript{0.05} & --0.57\textsubscript{0.03} & {--0.56}\textsubscript{0.03} & {0.48}\textsubscript{0.02} \\
Fluency & {3.7}\textsubscript{1.1} & {0.75}\textsubscript{0.03} & ---\phantom{----} & --0.59\textsubscript{0.04} & {--0.56}\textsubscript{0.02} & {0.51}\textsubscript{0.03} \\ 
Both & 3.8\textsubscript{1.0} & 0.83\textsubscript{0.03} & ---\phantom{----} & --0.63\textsubscript{0.04} & --0.61\textsubscript{0.03} & 0.53\textsubscript{0.03} \\
\bottomrule
\end{tabular}
}
\end{table}

Table~\ref{tab:human-eval-correlations} summarizes the human judgments.
\textit{Meaning} and \textit{fluency} have mean scores of 3.9 and 3.7, respectively, indicating a relatively high satisfaction with the outputs.
Both metrics are strongly correlated with each other (Spearman's $\rho=0.73$) and moderately correlated with the automatic metrics relative to the Standard German references ($0.48\leq|\rho|\leq0.59$).
The correlations are a bit higher when considering the 
mean of the \textit{meaning} and \textit{fluency} scores 
($0.53\leq|\rho|\leq0.63$), 
indicating that the automatic measures to some extent capture the interplay of these dimensions.

\subsubsection{Error analysis}
\label{sec:error-analysis}

We extend the analysis of differences between the dialectal and Standard German references in~\S\ref{sec:data-differences} with a word-level error analysis of the predictions produced by Whisper large-v3 (ignoring punctuation/\allowbreak{}casing).
We distinguish between three types of outcomes: \sameAsRef~the model's hypothesis is identical to the reference word, \validAlt~it presents a valid alternative,\footnote{As valid alternatives~(\validAlt), we allow orthographically valid spelling variation, alternative translations of dialect-specific words, words/phrases that are closer to the dialect but acceptable in (informal) Standard German, and translations that are more normalized than the Standard German references (while still faithful to the audio).
We encounter all of these types of alternatives, similarly to prior work that anecdotally remarks on Whisper normalizing dialectal inputs \cite{dolev-etal-2024-whisper, gorisch-schmidt-2024-evaluating} while sometimes still preserving non-standard expressions \cite{gorisch-schmidt-2024-evaluating}.} or \raisebox{-0.5pt}{\error}~it constitutes an error (a wrongly transcribed word, an insertion, or a deletion).

Table~\ref{tab:example} provides an example, and Table~\ref{tab:wordlevel-differences} summarizes the results.
Most of the words that are identical in the two references~(\qty{86}{\percent}) or that differ in terms of their pronunciation or morphology~(\qty{75}{\percent}) are transcribed correctly.
When encountering dialect-specific word choices, the model 
typically generates errors~(\qty{63}{\percent}).
Many of the grammatical constructions that differ in the transcripts are transcribed similarly to the dialectal reference -- whether this structure is also acceptable in (informal) Standard German depends on the constructions.

An additional analysis of the errors shows that relatively many of them (\qty{27}{\percent}) are due to the model not recognizing word boundaries properly.

\section{Conclusion}

We introduce \betthupferl{}, a challenging benchmark for German dialect-to-standard speech translation and dialect ASR.
Current state-of-the-art ASR models show a substantial performance gap between transcribing the dialectal and Standard German audios in this dataset.
Automatic metrics do not fully correlate with human judgments and sometimes indicate errors where a model instead produced a valid alternative transcription.
We hope that our dataset encourages future work on evaluating ASR systems on non-standard dialects, both relative to their standard-language references and to dialectal ones (cf.~\cite{nigmatulina-etal-2020-asr, aepli-etal-2023-benchmark}).

\section{Acknowledgements}
We thank Evangelos Ziogas and Franziska Schmidt for providing ASR quality judgments.
We also thank Siyao Peng, Ryan Soh-Eun Shim and Bolei Ma for feedback on paper drafts.
This research is supported by the ERC Consolidator Grant DIALECT 101043235.

\bibliographystyle{IEEEtran}
\bibliography{mybib}

\end{document}